\documentclass[runningheads]{llncs}

\usepackage{eccv}

\usepackage{eccvabbrv}

\usepackage{graphicx}
\usepackage{booktabs}

\usepackage[accsupp]{axessibility}  

\newcommand{\supp}{\text{supp}}

\usepackage{hyperref}

\usepackage{orcidlink}

\begin{document}

\title{\texttt{Control+Shift}: Generating Controllable Distribution Shifts} 


\author{Roy Friedman$^\star$ \orcidlink{0000-0001-5061-4084} \and
Rhea S Chowers$^\star$ \orcidlink{0009-0001-5066-5794}}

\authorrunning{Friedman and Chowers}

\institute{The Hebrew University of Jerusalem, Israel}

\def\thefootnote{$\star$}\footnotetext{These authors contributed equally to this work.}\def\thefootnote{\arabic{footnote}}

\maketitle

\begin{abstract}

    
    We propose a new method for generating realistic datasets with distribution shifts using any decoder-based generative model. Our approach systematically creates datasets with varying intensities of distribution shifts, facilitating a comprehensive analysis of model performance degradation.
    We then use these generated datasets to evaluate the performance of various commonly used networks and observe a consistent decline in performance with increasing shift intensity, even when the effect is almost perceptually unnoticeable to the human eye. We see this degradation even when using data augmentations. We also find that enlarging the training dataset beyond a certain point has no effect on the robustness and that stronger inductive biases increase robustness.
\end{abstract}

\section{Introduction}\label{sec:intro}

Data in the real world rarely follows the assumptions made in current training regimes, where the samples seen during deployment are assumed to be i.i.d. from the same distribution as the model was trained on~\cite{koh2021wilds,beery2018recognition,recht2018cifar,recht2019imagenet}. Instead, the test distribution can shift from the training distribution in unexpected manners. For instance, the focal lengths most frequently used to capture images may change over time or objects that did not exist at the time of training can suddenly appear in newly acquired data (e.g. new models of cars).

\begin{figure}
\begin{center}
    \includegraphics[width=\linewidth]{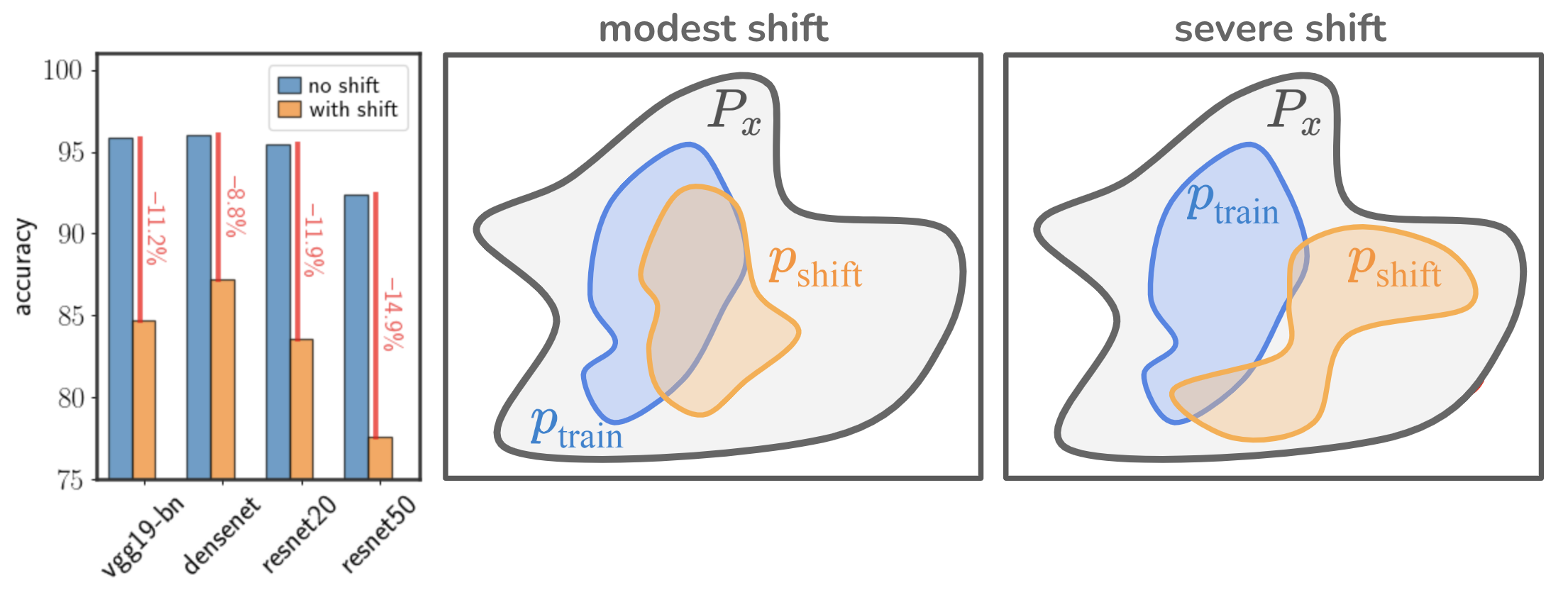}
\end{center}
  \caption{On the left, the effects for a specific distribution shift is displayed as the accuracy on the shifted distribution versus the accuracy on the training distribution. For this plot, the train and shifted distribution are from our ``extend shift'' dataset described in~\cref{sec:generated-data}.
  The center and right plots show a schematic for distribution shift in the real world, where the amount of shift can be either modest in some sense (center) or severe (right). In both cases, we expect the training distribution to be a subset of the full distribution in question and the distribution shift is a distinct subset from the training distribution.}
  \label{fig:dist-shift}

\end{figure}

While humans are mostly invariant to various distribution shifts, such as changes in lighting, angle, color and more, it has been shown that neural networks are often sensitive to such shifts. This is true even when using state of the art models~\cite{shifman2024lost,taori2020measuring}. An example of the effects of such a change can be seen in~\cref{fig:dist-shift} (left), where the accuracy of a model initially trained on one distribution drops significantly on a test set that is shifted from it - both test sets are from datasets we generated (see~\cref{sec:generated-data}). This large drop in accuracy occurs despite the fact that images from both distribution look perceptually equivalent to us.


In recent years, a large body of work has been devoted to categorizing~\cite{quinonero2022dataset,liu2021towards,wiles2021fine,geirhos2020shortcut}, benchmarking~\cite{hendrycks2019benchmarking,xiao2020noise,galil2023framework} and mitigating the degradation in performance caused by such shifts in the test distribution~\cite{arjovsky2019invariant,sagawa2019distributionally}. To this end, an impressive range of datasets were produced to better understand the impacts and behavior of distribution shift. These datasets generally fall into one of two categories. The first is unrealistic synthetic data that is easy to use and has a sense of the severity of the distribution shift (such as \cite{dsprites17,kim2018disentangling,bordes2023pug}). In these datasets the shift often has semantic meaning such as change in color or shape.
The other datasets are of realistic images undergoing distribution shift where not much is known about the properties or severity of the  shift (such as \cite{koh2021wilds}). These datasets practically create a discrepancy between controllability of the distribution shift and realism of the setting when seeking to test robustness of models to different shifts.

In this work, we aim to bridge these two regimes by creating datasets of realistic images that undergo distribution shifts of varying, measurable, intensities. As a first step, we model the shifted distribution as one that has a partial intersection with the training distribution, as seen in~\cref{fig:dist-shift} (center and right). Under this framework, we model the intensity of the shift as the amount to which the shifted and training distributions' supports overlap. In mild distribution shift the overlap is large (\cref{fig:dist-shift} center) while in severe distribution shift the overlap is small (\cref{fig:dist-shift} right).
Using the above definitions, we utilize the recent advances of generative models~\cite{ho2020denoising,sauer2022stylegan,karras2022elucidating} in order to generate realistic datasets of images with controllable intensities of distribution shift, different amounts of training data and multiple types of distribution shifts.

These new datasets enable careful investigation into what makes models more or less robust. We then test the performance of various classifiers on our newly generated datasets and gain some surprising insights.
First, training with augmentations increases robustness but models trained this way still suffer under distribution shift. Second, increasing the size of the dataset used for training is not enough to gain robustness when confronted with distribution shift. Moreover, convolutional networks tend to be more robust than other architectural choices on our image datasets. Finally, we empirically observe that the degradation of performance in many models is linear in a sense of distance from the support of the training distribution.

\section{Previous Work}

\subsection{Distribution Shift}
The fragility of classifiers to distribution shift is an active field of research and as such there are many different categorizations of distribution shifts~\cite{quinonero2022dataset}. At the broadest level, distribution shift can be divided into covariate shift, where the distribution of the observed data changes, and concept shift, where the labeling process changes. In this work we focus solely on covariate shift.

In the world of covariate shift, it has been shown that current recognition models are sensitive to changes in sub-populations that are underrepresented~\cite{koh2021wilds,beery2018recognition}, are fragile to small perturbations in images~\cite{hendrycks2019benchmarking,hendrycks2021natural,shifman2024lost}, incorrectly classify objects on a backgrounds they are not typically paired with~\cite{xiao2020noise,geirhos2020shortcut}, among other instabilities to changes in the test distribution. 

\subsection{Existing Benchmarks for Distribution Shift}
To better investigate the robustness of classifiers, many synthetic and realistic datasets have been proposed and utilized. Below we give a brief overview of these datasets.

\subsubsection{Synthetic Distribution Shift} 
The datasets that are easiest to manipulate in order to investigate specific types of distribution shift are those of synthetic data with tuneable coefficients of variation. The coefficients of variation can either have semantic meaning (type of object in the image) or not (such as added Gaussian noise). These include simple datasets of rendered geometric shapes~\cite{dsprites17,kim2018disentangling} or rendering of 3D objects of various colors, poses and backgrounds~\cite{chang2015shapenet,bordes2023pug}. Slightly more realistic datasets with controllable shifts introduce changes to standard benchmark datasets such as CIFAR10 and ImageNet~\cite{hendrycks2019benchmarking,xiao2020noise}.

While these datasets are incredibly useful for gaining an understanding of the behavior of classifiers in domains not seen during training, they are not realistic in the sense that they either do not have the natural variance seen in true images or include shifts that we wouldn't expect to observe in deployment.

\subsubsection{Real Data Undergoing Distribution Shift}
To understand how recognition models behave in a more realistic setting, a large number of datasets were gathered for distribution shifts closer to those observed in the wild. These datasets range from real images categorized to different domains and classes~\cite{he2021towards,zhang2023nico++,sagawa2019distributionally}, to a collection of datasets in the wild where it is known that some distribution shift has occurred~\cite{koh2021wilds}. In all of these, it is known that the test distributions differ from the train in some way, but there is no sense as to the strength of the shift from the training distribution. 

We are primarily concerned with creating a dataset of \emph{realistic images} that explicitly undergoes a distribution shift whose severity is both known and controllable. 

\subsection{Decoder-Based Generative Models}

Recent advances in generative models have made large steps towards the generation of realistic samples (e.g. \cite{ho2020denoising,sauer2022stylegan,rombach2021highresolution,karras2022elucidating}). Our aim in this work is to utilize the fact that these generated images are of such high quality in order to generate a dataset that contains a distribution shift. We will describe how decoder-based generative models (DBGMs) can be used in order to generate such data.

A DBGM with generator $G$ can be abstractly described as a mapping function $G:\mathcal{Z}\times\mathcal{Y}\rightarrow\mathcal{X}$, where $\mathcal{Z}$ is called the latent space, $\mathcal{Y}$ is the space of labels when the generative model is conditional, and $\mathcal{X}$ is the output space. To sample from the generative model, the latent space $z\in\mathcal{Z}$ is imbued with a prior distribution, almost ubiquitously the standard normal distribution. A DBGM is then trained so that $x=G(z,y)$ is a sample from the training distribution, as long as $z$ is a sample from the prior distribution.

We used score-based models (SBM)~\cite{song2020score,karras2022elucidating} to generate datasets and will use their probability flow formulation as per Song \etal ~\cite{song2020score} to ensure that the mapping function $G(z,y)$ is deterministic. We use SBMs due to the quality of the generated images and because they are relatively easy to use. However, in principle any DBGM can be used with our methodology in order to generate shifted datasets.

\section{Generating Realistic Distribution Shifts}

\subsection{A Simple Model for Distribution Shift}\label{sec:shift-definition}
\begin{figure}
\begin{center}
    \includegraphics[width=\linewidth]{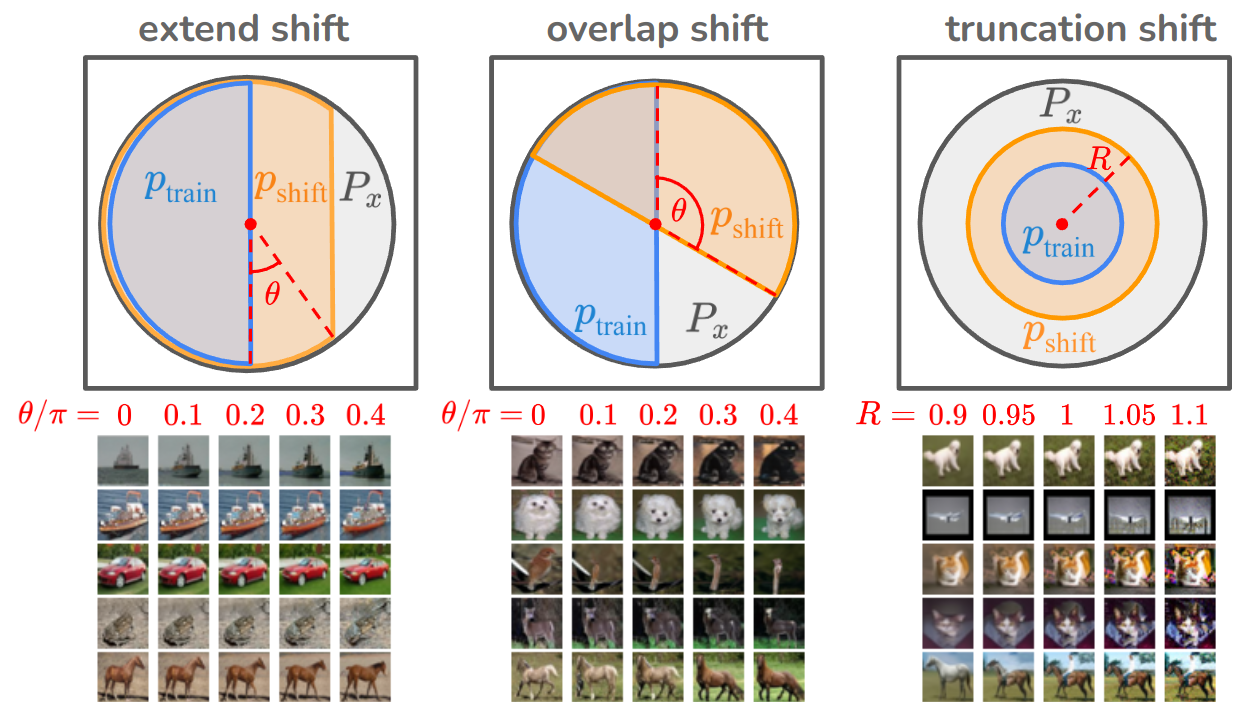}
\end{center}
  \caption{A schematic for the generation processes we explore (top) and images generated according to this procedure (bottom). In all types of shifts, the severity of the shift is controllable, while still resulting in CIFAR10-like images. The bottom half of the figure shows images generated with the same seed at different shift intensities, with the left-most column in each depicting the training set.}
  \label{fig:shift-types}

\end{figure}
%



Let $\mathcal{X}$ be the space of all data points from a set of classes, $p_\text{train}:\mathcal{X}\rightarrow\mathbb{R}_+$ be the distribution of training examples and $p_\text{shift}:\mathcal{X}\rightarrow\mathbb{R}_+$ be that of the distribution-shifted test samples. Given a distribution $p(x)$, we denote it's support as $\text{supp}(p)$. In case the support of $p(x)$ is full, we treat $\text{supp}(p)$ as the set that contains 99\% of the mass of the distribution.

We will assume that $\supp (p_\text{train}),\supp (p_\text{shift})\subseteq \supp (P_x)$ where $P_x$ is the distribution of the data we truly wish to classify. So, for instance, in the archetypal task of classifying between dogs and cats, $P_x$ is the distribution of all possible cat and dog images, while the training and test distributions only capture a small part of the overall distribution $P_x$. This could be either due to sampling bias, environmental factors or any previously mentioned cause of distribution shift. Throughout the following, we assume that the conditional distribution of the label $P_{y|x}$ remains constant, so that there is no concept shift.

In the following sections we will be interested in a very basic form of distribution shift, which we define in the following manner:

\begin{definition}[distribution shift and shift intensity]\label{def:dist-shift}
    Let $p_\text{train}$, $p_\text{shift}$ and $P_x$ be distributions whose supports are connected spaces such that:
    \begin{equation}
    \supp (p_\text{train}),\ \supp(p_\text{shift})\subseteq \supp (P_x)    
    \end{equation}
    We will say that there is a \emph{distribution shift} between $p_\text{shift}$ and $p_\text{train}$ if:
    \begin{equation}
        0 \le \mathcal{I} = 1-\frac{\left|\supp(p_\text{train})\cap \supp (p_\text{shift})\right|}{\left|\supp (p_\text{shift})\right|} < 1
    \end{equation}
    and we will call $\mathcal{I}$ the \emph{shift's intensity}.
\end{definition}


This definition, while not the standard~\cite{quinonero2022dataset,arjovsky2019invariant,wiles2021fine,kulinski2023towards,budhathoki2021did,cai2023diagnosing}, captures the simplest intuition behind distribution shift. Mainly, the shifted distribution $p_\text{shift}$ is a subset of a distribution we are interested in ($P_x$) but is fundamentally different than the training distribution $p_\text{train}$. This overly simplistic model of distribution shift also allows an intuitive sense of the impact of the distribution shift through the shift intensity, $\mathcal{I}$. 

The definition above does not take the distributions' density function into account, and is most relevant when they are close to uniform over their support. This doesn't have to be the case. For instance, the supports of $p_\text{train}$ and $p_\text{shift}$ could be the same but the densities are extremely concentrated in different regions of space, or the opposite case where their support are different except for a single region around which the densities are concentrated. However, most datasets (particularly of images) are extremely diverse and each data point is typically different from the others in the dataset. There might be repetitions of the same data point in a dataset, but the number of copies is usually negligible in proportion to the size of the full dataset. As such, we believe that this sense of distribution shift is appropriate for many datasets in the real world.

\begin{figure}
\begin{center}
    \includegraphics[width=\linewidth]{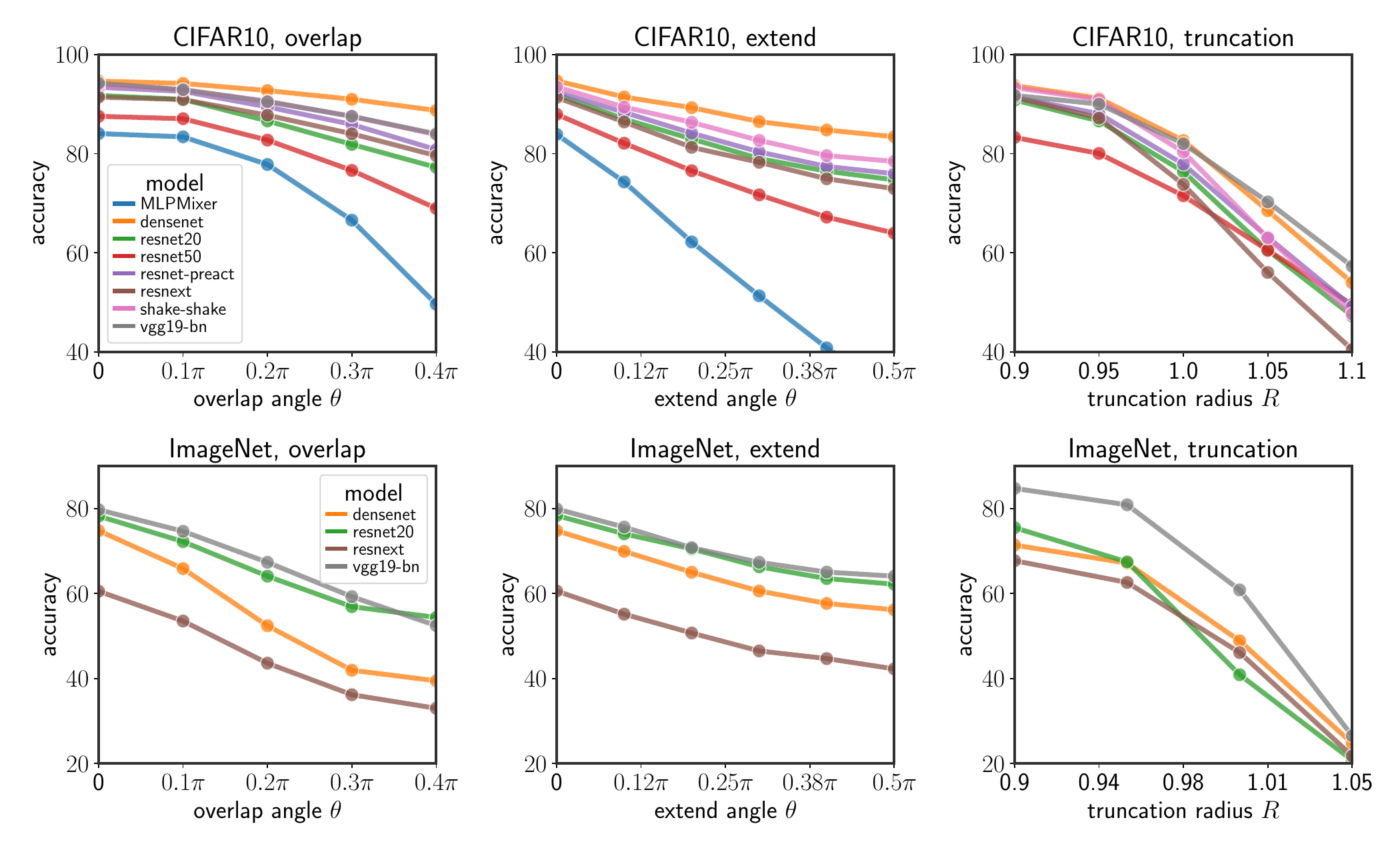}
\end{center}
  \caption{The accuracy of different models trained on the ``overlap'' (left), ``extend'' (center) and ``truncation'' (right) datasets and tested on the appropriate shifted distribution. 
  The top row shows results for the CIFAR10 variant of the distribution shifts and the bottom for the ImageNet generated shifts. In all cases, the trained models are not robust to the generated distribution shifts we present.
  The controlling parameters in this figure are the same as those shown in~\cref{fig:shift-types}, and the drop in accuracy is on the corresponding images.}
  \label{fig:performance}

\end{figure}

Typically, calculating the exact intensity of the shift will be impossible, since we are only supplied with a finite number of very high-dimensional samples from each distribution. We would still, however, like some sense of the strength of the shift. When needed, we will use the 1-NN distance between two datasets $D_\text{train}=\left\{y_i\sim p_\text{train}\right\}_{i=1}^N$ and $D_\text{shift}=\left\{x_i\sim p_\text{shift}\right\}_{i=1}^N$ defined in the following way:
\begin{equation}
    \text{1NN}_d(D_\text{train}, D_\text{shift})=\frac{1}{M}\sum_{x\in D_\text{shift}} \min_{y\in D_\text{train}} d(x,y)
\end{equation}
In words, the 1-NN distance is the average distance between each point in $D_\text{shift}$ to its nearest neighbor in $D_\text{train}$ under the distance function $d(x,y)$.

With these definitions in place, we develop a method for creating datasets with controllable shift intensities under the settings described above.

\subsection{Generation Procedure}\label{sec:gen-procedure}

Our goal is to generate data with controllable distribution shifts. To this end, we use decoder-based generative models (DBGMs) which map from a latent space $\mathcal{Z}$ with a known prior (usually a normal distribution) into the image space $\mathcal{X}$. From here on out, we will assume that the mapping defined by the DBGM $G(z,y)$ is a continuously differentiable, injective function. In other words each latent code $z\in\mathcal{Z}$ outputs a distinct image $x\in\mathcal{X}$ and a continuous change in $z$ implies a continuous change in $x$. 

To generate distribution shift under~\cref{def:dist-shift}, define two subsets from the prior distribution's support $\mathcal{S}_\text{train},\mathcal{S}_\text{shift}\subset \mathcal{Z}$ such that $\left|\mathcal{S}_\text{train}\cap \mathcal{S}_\text{shift}\right|<\left|\mathcal{S}_\text{shift}\right|$.

Having chosen the two sets $\mathcal{S}_\text{train}$ and $\mathcal{S}_\text{shift}$, let $U(\mathcal{S})$ define a uniform distribution over the set $\mathcal{S}$. A training dataset and a shifted test dataset, can now be constructed in the following manner:
\begin{align}
    \mathcal{D}_\text{train} &= \left\{x_i=G(z_i):z_i\sim U(\mathcal{S}_\text{train}):\right\}_{i=1}^N \\
    \mathcal{D}_\text{shift} &= \left\{x_i=G(z_i):z_i\sim U(\mathcal{S}_\text{shift})\right\}_{i=1}^M 
\end{align}

This framework allows for very simple control over the intensity of the shift as per~\cref{def:dist-shift}. To generate mild distribution shift, we choose $\mathcal{S}_\text{shift}$ so that it mostly overlaps with $\mathcal{S}_\text{train}$. For severe distribution shift, we choose $\mathcal{S}_\text{shift}$ with barely any overlapping support to $\mathcal{S}_\text{train}$. Because we assumed earlier that the DBGM uses an injective function to map back to samples, the relative overlap in support between the two sets transfers back into the sample space.  

\section{Experiments}\label{experiment_section}

\subsection{Generated Distribution Shifts}\label{sec:generated-data}
Using the procedure for generating distribution shift described in~\cref{sec:gen-procedure}, we generated datasets undergoing 3 distinct distribution shifts for both CIFAR10 and a subset of 10 classes from ImageNet. To generate this data, we used the probabilistic flow version of EDM~\cite{karras2022elucidating}, a type of score-based generator that allows for state-of-the-art image generation in CIFAR10 and a $64\times64$ version of ImageNet. The latent space of EDM is a standard normal with the same dimension as the generated data ($32\times32\times3$ for CIFAR10 and $64\times64\times3$ for ImageNet).

The types of distribution shifts we generated are (see Appendix~\ref{app:detailed-gen-instructions} for detailed descriptions of the generation process):
\begin{description}
    \item[Extend Shift:] a hemisphere of the unit hypersphere was chosen as the training distribution in this setting. The shifted distributions are then sampled from a larger range on the norm-1 ball in the latent space. The parameter controlling the shift intensity is the angle from the hemisphere's edge to that of the shifted distribution, as illustrated in~\cref{fig:shift-types} (left).
    \item[Overlap Shift:] in this setting, a hemisphere of the unit hypersphere is shifted across the latent space. The parameter controlling the shift intensity is the angle between the initial position of the hemisphere to that of the new position, as illustrated in~\cref{fig:shift-types} (center).
    \item[Truncation Shift:] in this setting, the training distribution is a standard normal with a variance smaller than 1. The shifted distributions are then Gaussians with increasing variances, as illustrated in~\cref{fig:shift-types} (right). The intensity of the shift is thus controlled by the norm of the latent code.
\end{description}
Notice that this idea can be extended to more types of shifts as any two arbitrary subsets of $\mathcal{Z}$ induce a distribution shift. The size of the datasets we created are the same as CIFAR10's train and test sets.

\begin{figure}
\begin{center}
    \includegraphics[width=\linewidth]{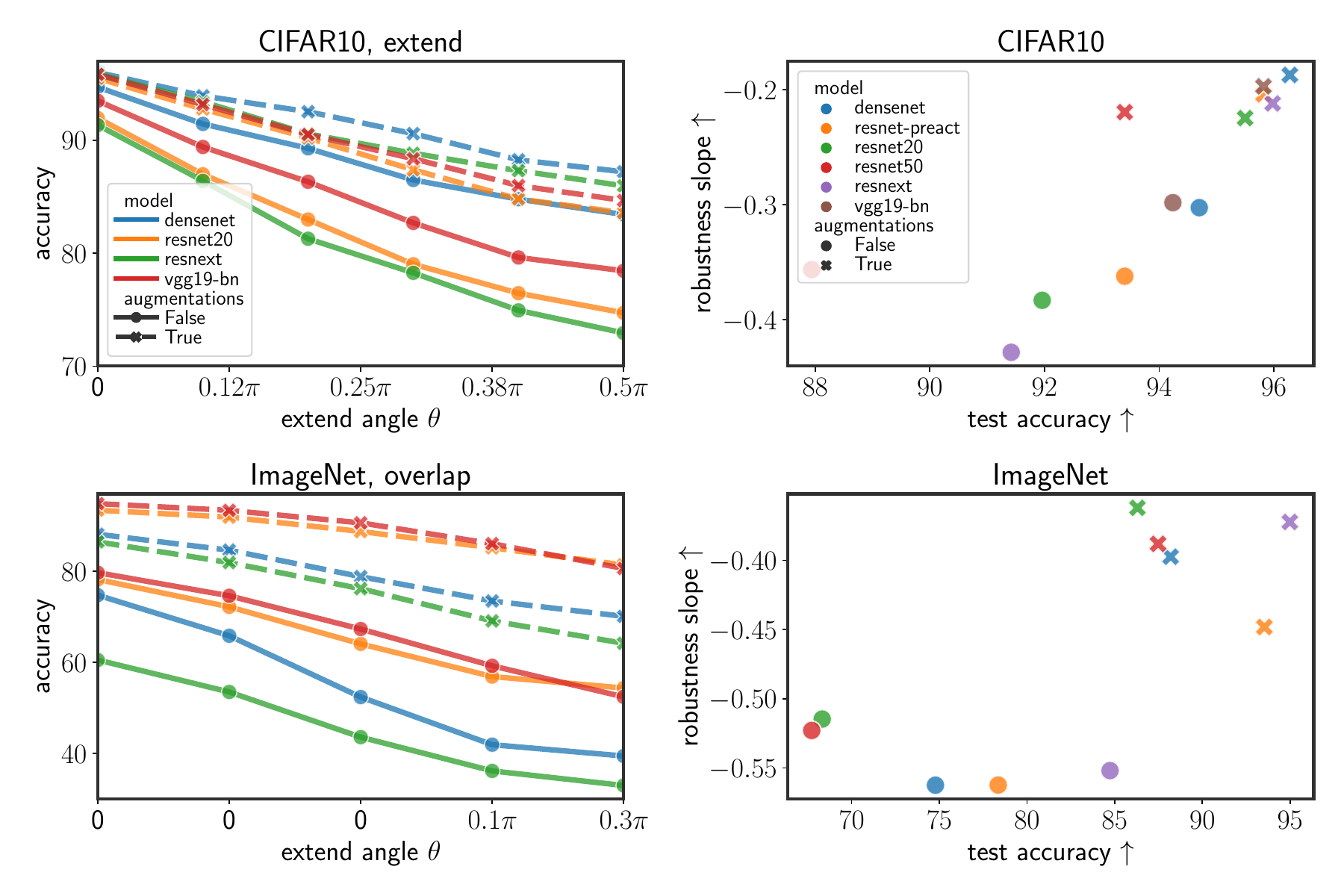}
\end{center}
  \caption{The accuracy as a function of shift on the ``extend'' dataset for both the CIFAR10 (top) and ImageNet (bottom) variants, without and with augmentation. Additionally, we show plots of the robustness slope as a function of the test accuracy of the models. In all cases, augmentation drastically helps with robustness.}
  \label{fig:aug-performance}

\end{figure}

As mentioned, \cref{fig:shift-types} includes schematics for the above types of distribution shifts, but also the images generated for each type. All images in the shifted distributions are generated with a constant seed, allowing us to visualize the change in the images that each shift type induces, as shown in the bottom half of~\cref{fig:shift-types} for CIFAR10 (see Appendix~\ref{app:more-images} for more generated images). For each shift type, the left-most column are images from the test set created in the training distribution, and the other columns show samples from the test sets for the shifted distributions. 

Even though the images don't seem drastically different as a function of the intensity of the shift, many classifiers' performance are severely degraded on these distribution shifts. This can be seen in~\cref{fig:performance}, where we trained classifiers on the training distribution of each of the shift types and then tested them on different intensities of shifts. Of the three types of distribution shift, the truncation shift seems to have the strongest effect. However, in all of these datasets the trained classifiers generalize poorly from the training distribution into the introduced shifts.

By choosing to use EDM to generate our distribution shifts, the generated datasets look perceptually very similar to CIFAR10 and ImageNet. We believe that using datasets such as these will allow a better investigation into the effects of distribution shift in settings close to those of the real world without the difficulty of collecting images with explicitly designed distribution shifts.

\subsection{Measuring Robustness}

Using our generated datasets, we explore the robustness of a range of classifiers commonly used in image classification~\cite{simonyan2014very,he2016deep,he2016identity,xie2017aggregated,huang2017densely,gastaldi2017shake,dosovitskiy2020image,tolstikhin2021mlp}. We are mainly interested in the drop in accuracy which we denote $\Delta$-accuracy, as a function of the shift intensity in either the latent space $\mathcal{Z}$ or the image space $\mathcal{X}$. To do so, we plot the $\Delta$-accuracy of models as either a function of 
the controlling parameter of the shift intensity in the latent space (overlap angle), or the 1-NN distance defined in~\cref{sec:shift-definition} as a proxy for measuring the shift in the image space. As we are interested in the rate of degradation, we look at the slope of said fit, which we call the \emph{robustness slope}. A robustness slope of 0 means the model is completely robust to our distribution shift, while a very negative slope implies that performance degrades quickly as a function of the 1-NN distance of the training distribution to the shifted distribution.

\begin{figure}
\begin{center}
    \includegraphics[width=\linewidth]{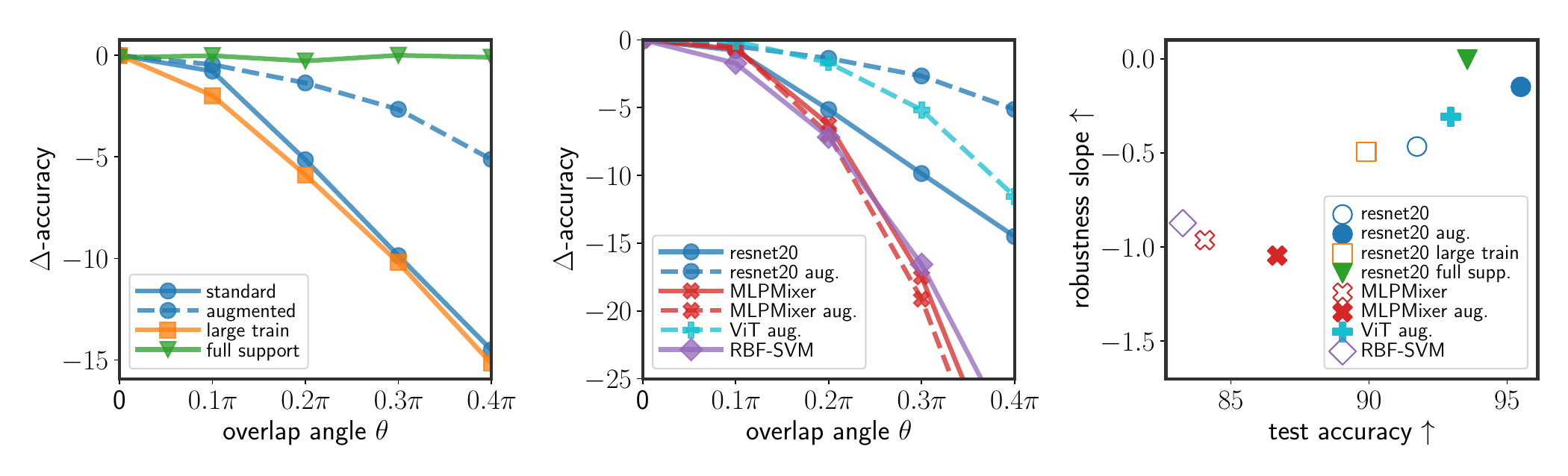}
\end{center}
  \caption{\textbf{Left}: $\Delta$-accuracy as a function of overlap angle in the ``overlap'' dataset for different sizes and supports of training distribution. \textbf{Center}: $\Delta$-accuracy as a function of overlap angle in the ``overlap'' dataset for models with different inductive biases. \textbf{Right}: the robustness slope as a function of test accuracy for all the models evaluated in the left and center plot.}
  \label{fig:large-pref}

\end{figure}

\subsubsection{Augmentations} 
Augmentations during training have become a standard protocol and are frequently mentioned as a way to add robustness to distribution shift (e.g.~\cite{geirhos2018imagenet,hendrycks2019augmix,yin2019fourier,goel2020model,rusak2021if,rebuffi2021data}). We test whether augmentations help with robustness in the datasets we generated as well by training models with and without RandAugment~\cite{cubuk2020randaugment} with the settings used by default. We find that adding augmentations does indeed help with robustness in many cases. An example for the ``extend shift'' dataset can be seen in~\cref{fig:aug-performance} (left). For all of the tested models, the baseline accuracy increased and the rate of degradation decreased. \cref{fig:aug-performance} (right) shows this more quantitatively - the robustness slope of all models increased (i.e. the models were more robust) when adding augmentations during training.

While adding augmentations do indeed increases the robustness of all of the tested models to our distribution shifts, it is important to note that there is still quite a sharp drop in accuracy when moving from the training dataset to the shifted ones even when augmenting, of around 10\%. In practice, using augmentations during training increases the support of the observed distribution during training.
This might serve as a possible explanation to why adding augmentations increased the robustness by decreasing the effective shift intensity observed by the tested models. 

\subsubsection{Dataset Size}
Another aspect of training that has previously been shown to help with robustness is the size and diversity of the training distribution. As we can generate training sets of any sizes using the procedure introduced in~\cref{sec:gen-procedure}, we use our framework to test this hypothesis as well. \cref{fig:large-pref} (left) shows the $\Delta$-accuracy of ResNet20~\cite{he2016deep} trained with different training datasets on our ``overlap shift'' dataset. In particular, training ResNet20 on twice as many samples (``large train'' in orange) has basically the same robustness as ResNet20 trained on the standard training set (``standard'' in blue). These results suggest that the factor that helps robustness is the support of the underlying training distribution, not simply the number of samples. To verify this fact, we train ResNet20 on images generated from all of the latent space of our generator; the performance of this model (``full support'' in green) is completely robust, unlike the model trained on twice as many samples.

\subsubsection{Inductive Biases}
The type of architecture used as the classifier can potentially have a large impacts on the classification robustness, as different types of models have different inductive biases. 
\cref{fig:large-pref} (center) shows the $\Delta$-accuracy on the ``overlap shift'' datasets for classifiers that range from weak to strong inductive biases: RBF-SVM, MLPMixer~\cite{tolstikhin2021mlp}, ViT~\cite{dosovitskiy2020image} and ResNet20. Our main takeaway from these experiments is that a strong inductive bias is crucial for better robustness in the face of distribution shift such as introduced by the ``overlap shift''.

\subsection{Robustness Over Distance}
\begin{figure}
\begin{center}
    \includegraphics[width=.5\linewidth]{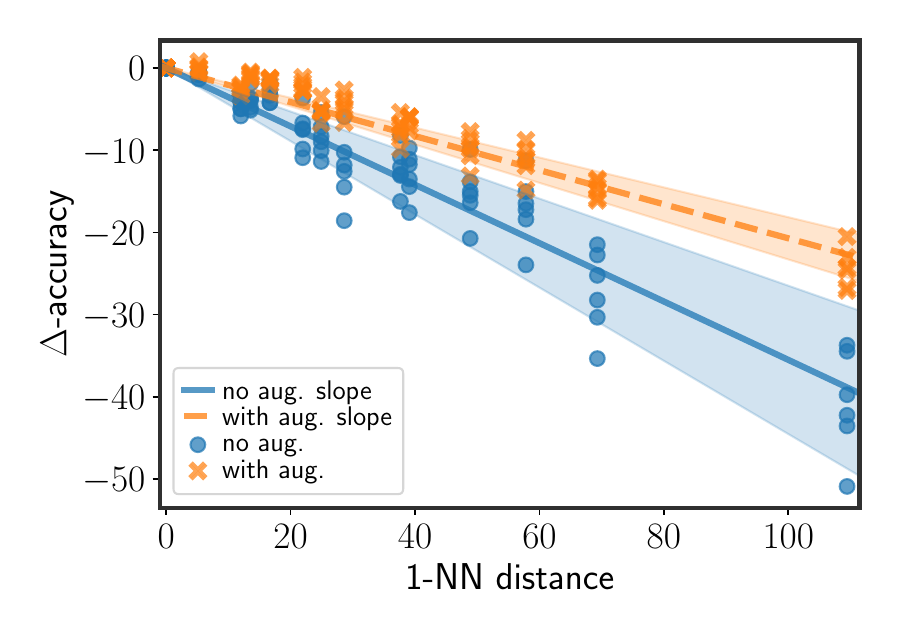}
\end{center}
  \caption{The $\Delta$-accuracy of all models trained as a function of 1-NN distance for the CIFAR10 variants.  Surprisingly, the $\Delta$-accuracy is linear in the 1-NN distance. The blue circles depict models trained without augmentation while the orange crosses are those trained with augmentation, which makes it clear that adding augmentations to the training procedure increases the robustness, but still doesn't completely solve the problem.}
  \label{fig:1nn-distance}

\end{figure}
To better understand the role of the intensity of the shift on the drop in accuracy, we plot the $\Delta$-accuracy as a function of the 1-NN distance defined in~\cref{sec:shift-definition} of the different generated datasets in~\cref{fig:1nn-distance}. We find that the drop in accuracy is mostly a linear function of the 1-NN distance between the shifted dataset and the training dataset. This result follows the surprising findings in~\cite{taori2020measuring} that show that the drop in accuracy from one dataset to another is consistent between many models.

Our results extend this by showing that the slope of the linear trend changes according to the robustness of the model. In~\cref{fig:1nn-distance}, we plot the slope for models trained without augmentation (solid blue line) and those trained with augmentations (dashed orange line). As we have shown previously, training with augmentations seems to increase robustness, and this is clearly apparent in these results as well - the slope of the degradation as a function of distance is smaller for models trained with augmentation than those trained without.

These results hint that quantifying the drop in accuracy incurred under distribution shift could be possible to some degree. Moreover, it shows that there is a strong connection between the classification rules learned by the models and a sense of distance from the training dataset. Such behavior is counter to what should be expected from robust classifiers; the pixel values in two images can be very distant, but the main objects very similar.

 
\section{Discussion}


The goal of classification, especially in the face of covariate shift, is to be both accurate \emph{and} robust. Current methods are extremely accurate on curated training datasets and their respective test sets, but are surprisingly fragile to small changes in the testing environment, some even imperceptible to humans. 

As a step towards gaining a better understanding of this behavior, we defined a simplistic but intuitive case of distribution shift meant to simulate the real world. In particular, any \emph{finite} dataset will always capture only a partial region of the support of the underlying distribution, and there is no reason to believe that any test samples are localized to the same portion of the support. Our framework for the generation of realistic data is meant to capture this setting with the explicit goal of measuring sensitivity to distribution shifts of different intensities.

All of the classifiers we trained and tested on our datasets were sensitive to distribution shift. Similar to previous works, we found that adding augmentations during training helps increase a model's robustness to distribution shift. However, even with augmentations the performance still suffered significantly in the presence of distribution shift, suggesting that these models haven't learned representations that can truly be called robust. Our experiments also showed that increasing the number of samples from the training distribution alone does not result in more robust models, a result that is in contrast to previous empirical observations.

In section \cref{experiment_section}, we showed that models with stronger inductive biases relevant to image recognition are relatively more robust to our distribution shifts. We believe that this insight could be a direction for future research - understanding the failures of current inductive biases relative to such imperceptible changes and fixing them by additional biases, instead of additional data.

\bibliographystyle{splncs04}
\bibliography{main}

\clearpage

\appendix

\section{Detailed Overview of Generation Procedure}\label{app:detailed-gen-instructions}

\cref{sec:generated-data} described three different types of distribution shift we were concerned with in this work. Following are the precise instructions for generating samples for each of the datasets.

For each of them, let $z\sim\mathcal{N}(0, I)$ be an i.i.d. draw from the prior of the latent space of the generative model used. We will describe how $\tilde{z}$, the latent code fed into the generative model, is produced. For the extend and overlap distribution shifts, two additional \emph{target} codes $t_1,t_2\in\mathcal{Z}$ have to be used, which are constant throughout all experiments. In our generation procedure, $t_1$ and $t_2$ are random draws from $\mathcal{N}(0, I)$ with a specific seed.

\begin{description}
    \item[Truncation Shift:] this distribution shift is the simplest to explain. Given a truncation radios $R$, the shifted version of $z$ is $\tilde{z}=Rz$.
    \item[Extend Shift:] the goal in this distribution shift is to cover growing regions of the hypersphere. To achieve this, we use spherical linear interpolation (slerp) between $z$ and $t_1$ of different amounts. The slerp between $z$ and $t_1$ with parameter $\tau$ is defined as:
    \begin{equation}
        \Omega = \arccos(z^Tt_1)\rightarrow \text{slerp}(z,t_1;\tau) = \frac{\sin\left((1-\tau)\Omega\right)}{\sin \Omega}\cdot z + \frac{\sin (\tau \Omega)}{\sin \Omega}\cdot t_1
    \end{equation}
    In our implementation, we always normalize $z$ and $t_1$ to be on the norm-1 hypersphere, so that $\text{slerp}(z,t_1;\tau)$ is also on the same hypersphere. Given the extend angle $\theta$, the shifted version of $z$ is:
    \begin{equation}
        \tilde{z}=\text{slerp}\left(\frac{z}{\|z\|}, \frac{t_1}{\|t_1\|}\ ;\ \frac{\theta + \pi/2}{\pi}\right)
    \end{equation}
    In this way, when $\theta=0$ the interpolation parameter is equal to a half, and $\tilde{z}$ is part of the hemisphere defined by $t_1$. When $\theta =\pi/2$, the slerp returns the normalized version of $z$.
    \item[Overlap Shift:] this type of distribution shift is similar to the extend shift described above, only this time the size of the support of the defined distribution doesn't change. Given the parameter $\theta$, calculate the following vector:
    \begin{equation}
        \tilde{t}=\text{slerp}\left(\frac{t_1}{\|t_1\|}, \frac{t_2}{\|t_2\|}\ ;\ \frac{2\theta}{\pi}\right)
    \end{equation}
    and return:
    \begin{equation}
        \tilde{z} = \text{slerp}\left(\frac{z}{\|z\|},\tilde{t}\ ;\ 1/2\right)
    \end{equation}
\end{description}

\section{More Images}\label{app:more-images}

\begin{figure}[hbt!]
\begin{center}
    \includegraphics[width=\linewidth]{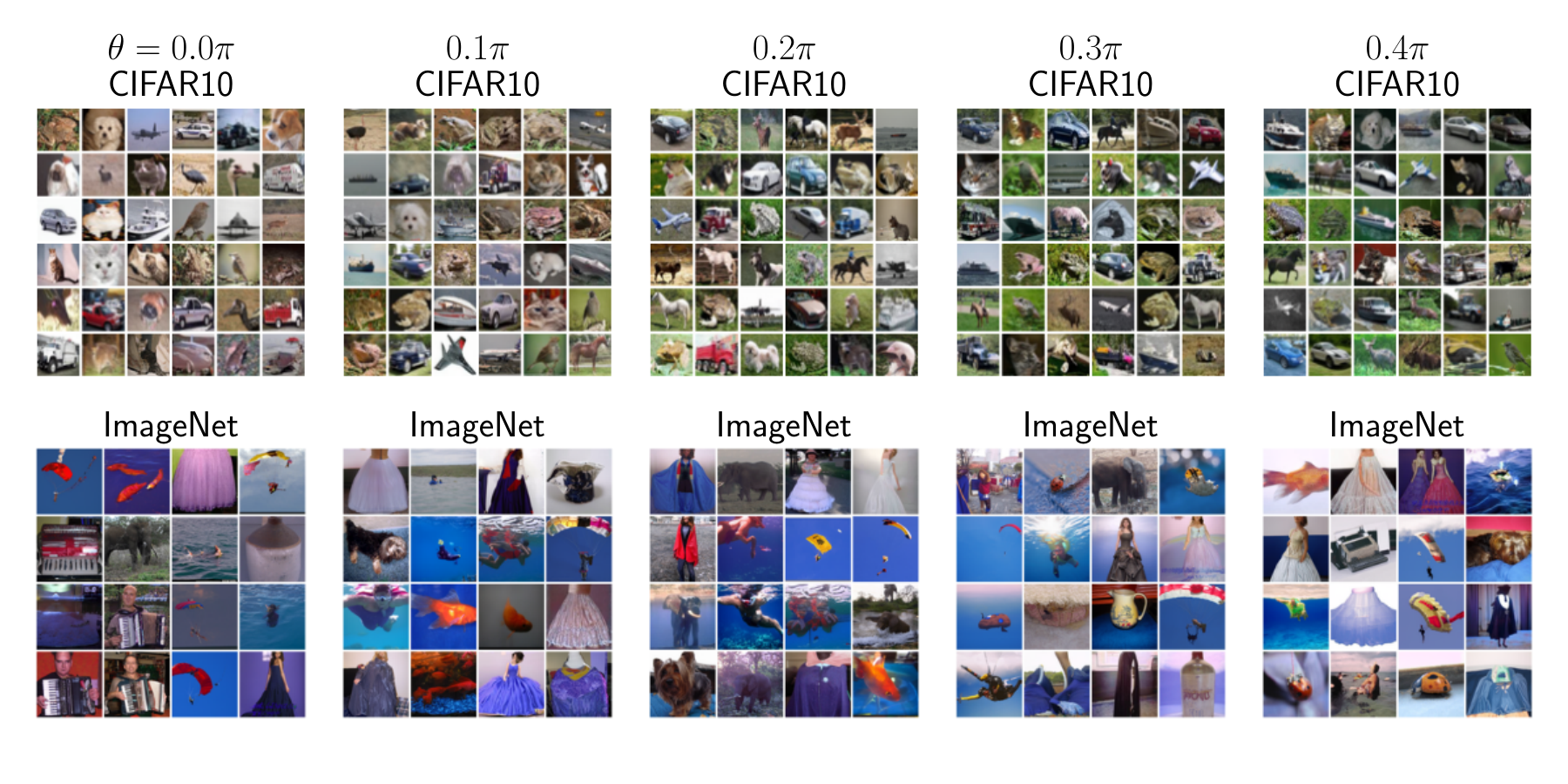}
\end{center}
  \caption{Randomly sampled images from the ``overlap shift'' dataset for both CIFAR10 and ImageNet.}
\end{figure}

\begin{figure}[hbt!]
\begin{center}
    \includegraphics[width=\linewidth]{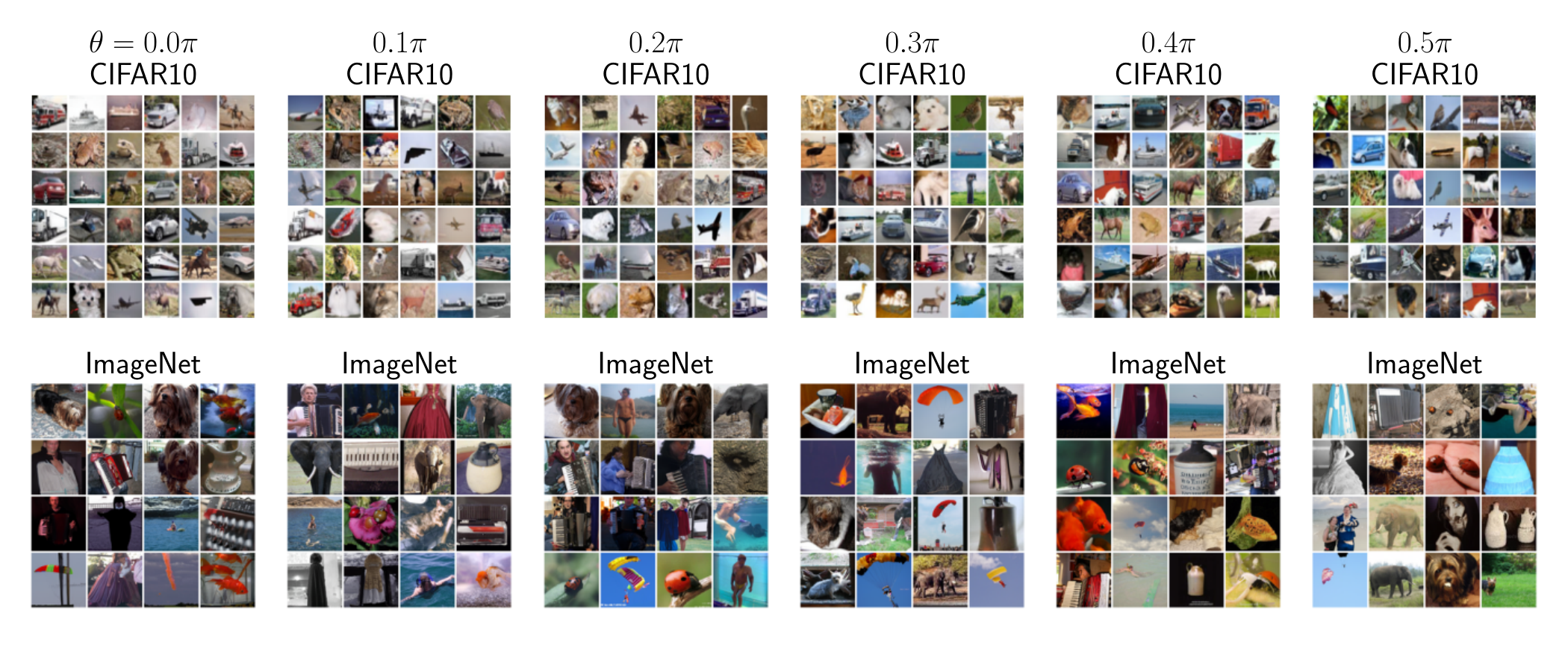}
\end{center}
  \caption{Randomly sampled images from the ``extend shift'' dataset for both CIFAR10 and ImageNet.}
\end{figure}

\begin{figure}[hbt!]
\begin{center}
    \includegraphics[width=\linewidth]{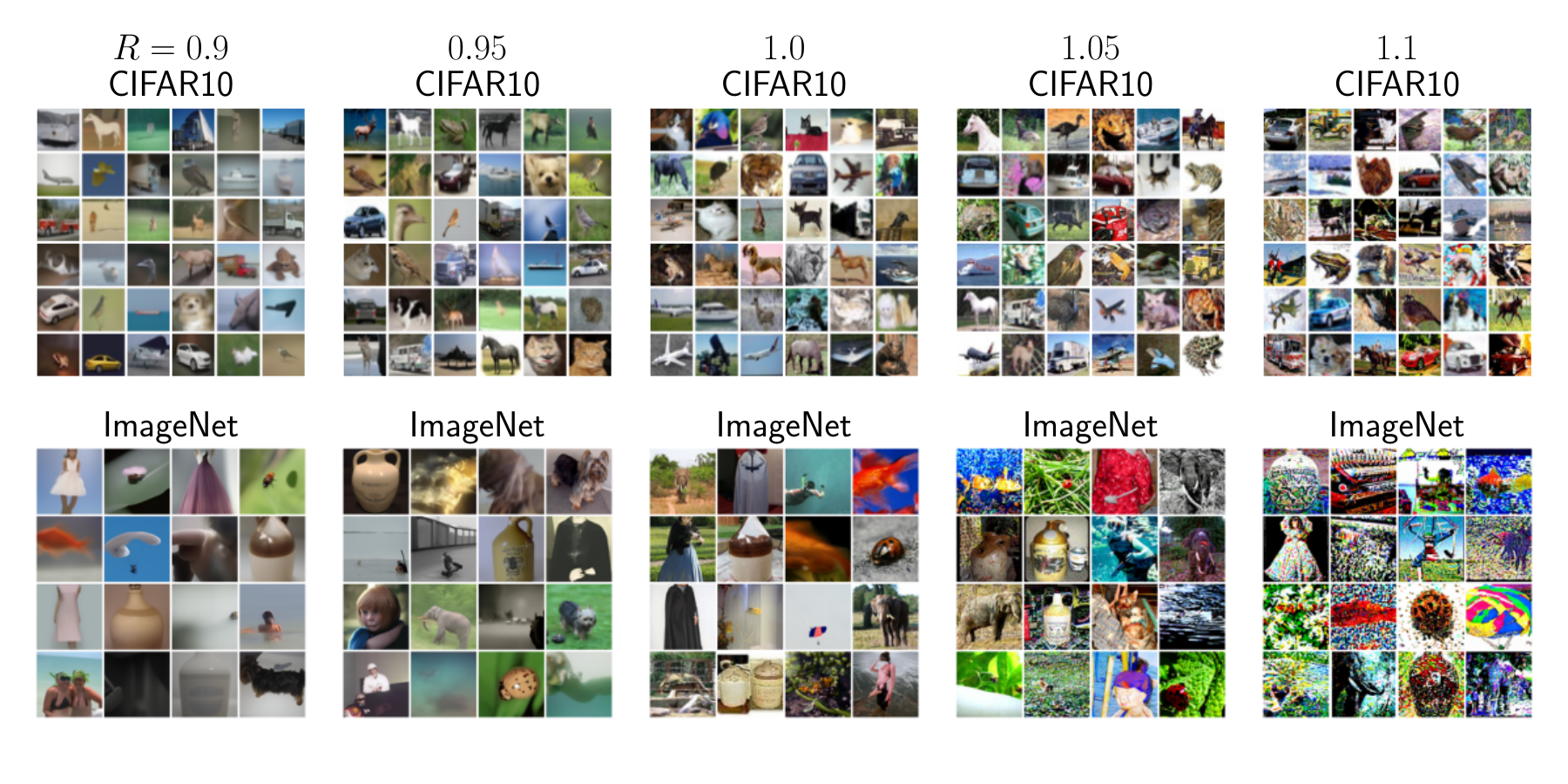}
\end{center}
  \caption{Randomly sampled images from the ``truncation shift'' dataset for both CIFAR10 and ImageNet. In the ImageNet experiments we typically don't use the $R=1.1$ images due to the strong generated artifacts.}
\end{figure}

\end{document}